\documentclass{article}

\newcommand{\calD}{\mathcal{D}}

\newcommand{\given}{\;|\;}

\newcommand{\thetassl}{\theta_\text{SSL}}
\newcommand{\thetaft}{\theta_\text{FT}}
\newcommand{\dssl}{\calD_\text{SSL}}
\newcommand{\dft}{\calD_\text{FT}}

\usepackage{amsmath, amsthm, amssymb}

\usepackage{expl3}  %
\ExplSyntaxOn
  _to_str:N
\ExplSyntaxOff
\usepackage{ifthen}
\usepackage{pgffor}

\usepackage[acronym,smallcaps,nowarn,section,nonumberlist]{glossaries}
\glsdisablehyper  %
\makeglossaries

\setacronymstyle{long-short}

\newacronym{auc}{AUC}{Area Under the Curve}

\newacronym{bert}{BERT}{Bidirectional Encoder Representations from Transformers}
\newacronym{bow}{BOW}{Bag of Words}
\newacronym{boe}{BOE}{Bag of Embeddings}
\newacronym{borep}{BOREP}{Bag of Random Embedding Projections}
\newacronym{bn}{BN}{Bayesian Network}

\newacronym{cdf}{CDF}{Cumulative Distribution Function}
\newacronym{cl}{CL}{Continual Learning}
\newacronym{cnn}{CNN}{Convolutional Neural Network}
\newacronym{cka}{CKA}{Centered Kernel Alignment}
\newacronym{cvae}{CVAE}{Conditional Variational Auto-Encoder}

\newacronym{dag}{DAG}{Directed Acyclic Graph}
\newacronym{dgm}{DGM}{Directed Graphical Model}
\newacronym{dnn}{DNN}{Deep Neural Network}

\newacronym{ecfp4}{ECFP4}{Extended Connectivity Fingerprint}
\newacronym{ehr}{EHR}{Electronic Health Record}
\newacronym{erm}{ERM}{empirical risk minimization}
\newacronym{esn}{ESN}{Echo State Network}
\newacronym{ewc}{EWC}{Elastic Weight Consolidation}

\newacronym{fft}{FFT}{Fast Fourier Transform}
\newacronym{fs}{FS}{FastSent}
\newacronym{ft}{FT}{Fourier Transform}

\newacronym{gat}{GAT}{Graph Attention Network}
\newacronym{gcn}{GCN}{Graph Convolutional Network}
\newacronym{gdl}{GDL}{Geometric Deep Learning}
\newacronym{ggnn}{GGNN}{Gated Graph Neural Network}
\newacronym{glove}{GloVe}{Global Vectors for Word Representation}
\newacronym{gnn}{GNN}{Graph Neural Network}
\newacronym{gru}{GRU}{Gated Recurrent Unit}

\newacronym{hmm}{HMM}{Hidden Markov Model}

\newacronym{idf}{IDF}{Inverse Document Frequency}
\newacronym{ig}{IG}{Information Gain}
\newacronym{isan}{ISAN}{Input Switched Affine Network}

\newacronym{jtt}{JTT}{Just Train Twice}
\newacronym{jsd}{JSD}{Jensen-Shannon Divergence}

\newacronym{kld}{KLD}{Kullback-Leibler Divergence}
\newacronym{knn}{KNN}{$K$-Nearest Neighbour}

\newacronym{lhs}{L.H.S.}{left hand side}
\newacronym{lstm}{LSTM}{Long Short Term Memory Network}

\newacronym{map}{MAP}{Maximum a Posteriori}
\newacronym{mc}{MC}{Monte Carlo}
\newacronym{mdl}{MDL}{Minimum Description Length}
\newacronym{ml}{ML}{Machine Learning}
\newacronym{mlp}{MLP}{Multi-Layer Perceptron}
\newacronym{mle}{MLE}{Maximum Likelihood Estimation}
\newacronym{mnist}{MNIST}{Modified National Institute of Standards and Technology}

\newacronym{nlp}{NLP}{Natural Language Processing}
\newacronym{nn}{NN}{Neural Network}
\newacronym{nll}{NLL}{Negative Log Likelihood}

\newacronym{ode}{ODE}{Ordinary Differential Equation}
\newacronym{oh}{OH}{One-Hot}

\newacronym{pdf}{PDF}{Probability Density Function}
\newacronym{pgm}{PGM}{Probabilistic Graphical Model}
\newacronym{fim}{FIM}{Fisher Information Matrix}
\newacronym{psd}{PSD}{positive semi-definite}

\newacronym{rwy}{RWY}{re-weighting mini-batches}
\newacronym{relu}{ReLU}{Rectified Linear Unit}
\newacronym{rdf}{RDF}{Resource Description Framework}
\newacronym{rgb}{RGB}{Red Green Blue}
\newacronym{rgc}{RGC}{Relational Graph Convolution}
\newacronym{rgat}{RGAT}{Relational Graph Attention Network}
\newacronym{rgcn}{RGCN}{Relational Graph Convolutional Network}
\newacronym{rhs}{R.H.S.}{Right Hand Side}
\newacronym{rnn}{RNN}{Recurrent Neural Network}
\newacronym{roc}{ROC}{Receiver Operating Characteristic}

\newacronym{suby}{SUBY}{subsampling large classes}
\newacronym{sgd}{SGD}{Stochastic Gradient Descent}
\newacronym{sg}{SG}{SkipGram}
\newacronym{simclr}{SimCLR}{Simple framework for Contrastive Learning of visual Representations}
\newacronym{ssl}{SSL}{Self-supervised representation learning}
\newacronym{st}{ST}{SkipThought}
\newacronym{sts}{STS}{Semantic Textual Similarity}

\newacronym{termfreq}{TF}{Term Frequency}
\newacronym{tfidf}{TF-IDF}{Term Frequency-Inverse Document Frequency}
\newacronym[plural=TPPs, descriptionplural={Temporal Point Processes}]{tpp}{TPP}{Temporal Point Process}

\newacronym{vae}{VAE}{Variational Auto-Encoder}
\newacronym{vit}{VIT}{Vision Transformer}

\newacronym{wga}{WGA}{Worst Subgroup Accuracy}
\newacronym{wirgat}{WIRGAT}{Within-Relation Graph Attention}
\newacronym{wl}{WL}{Weisfeiler-Lehman}

\usepackage{bm}
\usepackage[T1]{fontenc}  %
\usepackage[type1]{libertine}
\usepackage[libertine]{newtxmath}
\usepackage{mathtools}
\usepackage{caption}
\usepackage{subcaption}
\usepackage{multirow}
\usepackage{booktabs}

\DeclareRobustCommand{\mathup}[1]{\begingroup\changegreek\mathrm{#1}\endgroup}
\DeclareRobustCommand{\mathbfup}[1]{\begingroup\changegreekbf\mathbf{#1}\endgroup}
\DeclareRobustCommand{\mathbit}[1]{\bm{\mathit{#1}}}

\DeclareMathAlphabet{\mathsfit}{\encodingdefault}{\sfdefault}{m}{sl}
\SetMathAlphabet{\mathsfit}{bold}{\encodingdefault}{\sfdefault}{bx}{n}
\newcommand{\tens}[1]{\bm{\mathsfit{#1}}}

\newcommand{\constantvector}{\bm}               %

\newcommand{\constantmatrix}{\bm}               %
\newcommand{\constantmatrixgreek}{\mathbit}

\newcommand{\randomscalar}{\textnormal}         %
\newcommand{\randomscalargreek}{\mathup}

\newcommand{\randomvector}{\mathbf}             %
\newcommand{\randomvectorgreek}{\mathbfup}

\newcommand{\randommatrix}{\mathbf}             %
\newcommand{\randommatrixgreek}{\mathbfup}

\newcommand{\graphstyle}{\mathcal}              %

\newcommand{\tensorstyle}{\tens}                %

\newcommand{\setstyle}{\mathbb}                %

\def\alphabet{a,b,c,d,e,f,g,h,i,j,k,l,m,n,o,p,q,r,s,t,u,v,w,x,y,z}
\def\Alphabet{A,B,C,D,E,F,G,H,I,J,K,L,M,M,O,P,Q,R,S,T,U,V,W,X,Y,Z}
\def\greekalphabet{alpha,beta,gamma,delta,epsilon,varepsilon,zeta,eta,theta,vartheta,iota,kappa,varkappa,lambda,mu,nu,xi,pi,varpi,rho,varrho,sigma,varsigma,tau,upsilon,phi,varphi,chi,psi,omega}
\def\GreekAlphabet{Gamma,Delta,Theta,Lambda,Xi,Pi,Sigma,Upsilon,Phi,Psi,Omega}

\makeatletter
\def\changegreek{\@for\next:=\greekalphabet
	\do{\expandafter\let\csname\next\expandafter\endcsname\csname\next up\endcsname}}
\def\changegreekbf{\@for\next:=\greekalphabet
	\do{\expandafter\def\csname\next\expandafter\endcsname\expandafter{%
			\expandafter\bm\expandafter{\csname\next up\endcsname}}}}
\makeatother

\foreach \x in \alphabet {
	\expandafter\xdef\csname v\x\endcsname{\noexpand\ensuremath{\noexpand\constantvector{\x}}}

	\expandafter\xdef\csname ev\x\endcsname{\noexpand\ensuremath{\noexpand\x}}

	\ifthenelse{\equal{\x}{m}}{}{
		\expandafter\xdef\csname r\x\endcsname{\noexpand\ensuremath{\noexpand\randomscalar{\x}}}
	}

	\expandafter\xdef\csname rv\x\endcsname{\noexpand\ensuremath{\noexpand\randomvector{\x}}}
}

\foreach \x in \greekalphabet {
	\expandafter\xdef\csname v\x\endcsname{\noexpand\ensuremath{\noexpand\constantvector{\csname \x\endcsname}}}

	\expandafter\xdef\csname ev\x\endcsname{\noexpand\ensuremath{\noexpand{\csname \x \endcsname}}}

	\expandafter\xdef\csname r\x\endcsname{\noexpand\ensuremath{\noexpand\randomscalargreek{\csname \x\endcsname}}}

	\expandafter\xdef\csname rv\x\endcsname{\noexpand\ensuremath{\noexpand\randomvectorgreek{\csname \x\endcsname}}}
}

\foreach \x in \Alphabet {
	\expandafter\xdef\csname m\x\endcsname{\noexpand\ensuremath{\noexpand\constantmatrix{\x}}}

	\expandafter\xdef\csname em\x\endcsname{\noexpand\ensuremath{\noexpand\x}}

	\expandafter\xdef\csname rm\x\endcsname{\noexpand\ensuremath{\noexpand\randommatrix{\x}}}

	\expandafter\xdef\csname t\x\endcsname{\noexpand\ensuremath{\noexpand\tensorstyle{\x}}}

	\expandafter\xdef\csname g\x\endcsname{\noexpand\ensuremath{\noexpand\graphstyle{\x}}}

	\ifthenelse{\equal{\x}{E}}{}{
		\expandafter\xdef\csname s\x\endcsname{\noexpand\ensuremath{\noexpand\setstyle{\x}}}
	}

}

\foreach \x in \GreekAlphabet {
	\expandafter\xdef\csname m\x\endcsname{\noexpand\ensuremath{\noexpand\constantmatrixgreek{\csname \x\endcsname}}}

	\expandafter\xdef\csname rm\x\endcsname{\noexpand\ensuremath{\noexpand\randommatrixgreek{\csname \x\endcsname}}}
}

\DeclareMathOperator*{\argmax}{arg\,max}

\newcommand{\E}{\mathbb{E}}

\usepackage[final]{templates/neurips2022/neurips_2022}

\usepackage{url}

\title{Elastic Weight Consolidation Improves the Robustness of Self-Supervised Learning Methods under Transfer}

\author{%
    Andrius Ovsianas$^1$\thanks{Work done during Apple internship.}\quad Jason Ramapuram$^{2}$\quad Dan Busbridge$^{2}$ \\ \textbf{Eeshan Gunesh Dhekane}$^{2}$ \qquad \textbf{Russ Webb}$^{2}$ \\
	University of Cambridge$^1$, Apple$^2$ \\
	\texttt{ao464@cam.ac.uk} \\
	\texttt{\{jramapuram, dbusbridge, eeshan, rwebb\}@apple.com}
}

\begin{document}
\maketitle
\begin{abstract}
\vspace{-1em}\gls{ssl} methods provide an effective label-free initial condition for
fine-tuning downstream tasks. However, in numerous realistic scenarios, the
downstream task might be biased with respect to the target label distribution.
This in turn moves the learned fine-tuned model posterior away from the initial
(label) bias-free self-supervised model posterior. In this work, we re-interpret
\gls{ssl} fine-tuning under the lens of Bayesian continual learning and consider
regularization through the \gls{ewc} framework. We demonstrate that
self-regularization against an initial \gls{ssl} backbone improves worst
sub-group performance in Waterbirds by 5\% and
Celeb-A by 2\% when using the ViT-B/16 architecture. Furthermore, to help simplify the
use of \gls{ewc} with \gls{ssl}, we pre-compute and publicly release the \gls{fim}, evaluated with 10,000 ImageNet-1K variates evaluated on large modern \gls{ssl} architectures including
ViT-B/16 and ResNet50 trained with DINO.
\end{abstract}

\setcounter{footnote}{2}
\section{Introduction}
\label{sec:introduction}

Self-supervised learning (SSL) methods for learning representations have recently gained
popularity within the deep learning community, bridging the gap with
supervised discriminative methods in vision
\citep{DBLP:journals/corr/abs-2104-14294,DBLP:journals/corr/abs-2103-01988,DBLP:conf/nips/GrillSATRBDPGAP20,DBLP:conf/nips/ChenKSNH20,DBLP:conf/icml/ChenK0H20}.
While representations learned via SSL methods are free from label induced bias
\citep{DBLP:journals/corr/abs-2103-01988}, this can change during the process of
fine-tuning to a downstream task.

In the supervised regime \cite{DBLP:journals/corr/abs-1911-08731} showed
that models trained with \gls{erm} tend to be biased towards label
population distributions that are disproportionately represented within the
training dataset. Our objective with this work is to mitigate this drift through
the use of Bayesian continual learning where we investigate regularizing downstream
tasks towards their robust initial representation produced by the SSL
pre-training procedure \citep{DBLP:journals/corr/abs-2103-01988}.

We consider regularizers based on the Fisher Information Matrix (\gls{fim}),
which constrain the model parameters towards their initial SSL values, as in
\gls{cl} techniques, such as Elastic Weight Consolidation (\gls{ewc})
\citep{kirkpatrick2017overcoming}. \gls{cl} is a rich sub-field of machine
learning that seeks to minimize the effect of catastrophic forgetting
\citep{mccloskey1989catastrophic} -- the phenomenon where models trained in a
sequential manner tend to become biased towards the latest observed distribution.
To use \gls{ewc}, we compute the \gls{fim} for DINO \citep{DBLP:journals/corr/abs-2104-14294}
models pre-trained with ViT-B/16 \citep{DBLP:conf/iclr/DosovitskiyB0WZ21} and
ResNet50 \citep{DBLP:conf/cvpr/HeZRS16} architectures on a subset of ImageNet1k
\citep{DBLP:conf/cvpr/DengDSLL009} images\footnote{\gls{fim} available
at https://coming\_soon}.

We validate the accuracy of the \gls{fim} by analyzing reverse transfer
performance from CIFAR10 fine-tuning. We observe that the
\gls{fim} for the ViT-B/16 is poorly conditioned, making it impossible to fully
recover the SSL performance on ImageNet1k, and propose a method to alleviate
this. Finally, we show that with ViT-B/16 this regularization can be used
to improve worst group accuracy on Waterbirds \citep{DBLP:journals/corr/abs-1911-08731} and
Celeb-A \citep{DBLP:conf/iccv/LiuLWT15} by 5\% and
2\% respectively.

\section{Method}
\label{sec:methodology}

To keep fine-tuned representations close to their initial \gls{ssl} values we
consider techniques used in \gls{cl}, in particular \gls{ewc}
\citep{kirkpatrick2017overcoming}, and treat the SSL pre-training and fine-tuning
tasks as two distinct sequential tasks. To overcome catastrophic forgetting
\citep{mccloskey1989catastrophic}, where models trained with stochastic
gradient descent become biased towards the latest task distribution, EWC
regularizes model parameters towards their optimal values on previous tasks.
The regularization uses the Fisher Information Matrix (FIM) and is based on the Laplace approximation \citep{laplace1986memoires, DBLP:conf/nips/SmolaVE03}.
An intuitive way to look at this regularization is through the
lens of online Bayesian inference. In particular, we consider a given SSL model
as a statistical Bayesian model $p_\theta(y \given x)$ with prior $p(\theta)$.
Given two datasets, $\dssl$ and $\dft$, observed one after another, our
objective is to estimate parameters $\theta$. While the full posterior
distribution $p(\theta \given \dssl, \dft)$ might be intractable, a point
estimate can be computed using Laplace's approximation.

First, the posterior with respect to the SSL task, $p(\theta \given \dssl)$,
is approximated with a Normal distribution using a Taylor's expansion as shown in Eq \eqref{eqn:taylorP}:
\begin{align}
  \log p(\theta \given \dssl) \approx  - \frac{1}{2} (\theta - \thetassl)^\top H(\thetassl) (\theta - \thetassl) = - \frac{1}{2} \|\theta - \thetassl\|_{H(\thetassl)}^2. \label{eqn:taylorP}
\end{align}
Here $\thetassl$ is the \gls{map} estimate of $\log p(\theta \given \dssl)$
and $H(\thetassl)$
is the Hessian. A point estimate to $p(\theta \given \dssl, \dft)$ can then
be derived via Bayes rule:
\begin{align}
    \thetaft = \argmax_\theta \ \log p(\theta \given \dssl, \dft) \approx \argmax_\theta \ \log p(\dft \given \theta) - \frac{1}{2} \|\theta - \thetassl\|_{H(\thetassl)}^2. \label{eqn:pointEstimate}
\end{align}
Since the Hessian is quadratic in the number of model parameters, it is impractical to
store it for anything besides small models and therefore it is common to rely on approximations \citep{DBLP:journals/neco/Amari98,DBLP:conf/icml/MartensG15}. Here we
consider the diagonal \gls{fim} as in \gls{ewc}\footnote{Most practical implementations of \gls{ewc} use the empirical \gls{fim} due to the relaxed computation complexity of the calculation.
  This has been criticized in the context of natural gradient descent~\citep{DBLP:conf/nips/KunstnerHB19}, where the authors argue that it produces a biased estimate. Here, we compute the \gls{fim} by explicitly enumerating $E_{y|x}$, where $y|x$ in Eq \eqref{eqn:fim} is distributed as a conditional Categorical and is sampled from the model distribution.} \citep{kirkpatrick2017overcoming}:
\begin{align}
    F(\theta) = \frac{1}{N} \sum_{i=1}^N\E_{y \sim p_{\theta}(\cdot \given x_i)}\left[\left\{ \nabla_\theta \log p_{\theta}(y \given x_i) \right\}^2\right], \label{eqn:fim}
\end{align}
which has been used in prior work as an approximation to $H(\thetassl) \approx F(\theta)$~\citep{DBLP:conf/nips/DaxbergerKIEBH21}.
As $F(\theta)$ is guaranteed to be \gls{psd}, unlike $H(\thetassl)$ which is
guaranteed to be \gls{psd} only when $\thetassl$ is an exact \gls{map} estimate,
this leads to a better objective, which we use:
\begin{align}
    \thetaft = \argmax_\theta\  \log p(\dft \given \theta) - \lambda \|\theta - \thetassl\|_{F(\thetassl)}^2, \label{eqn:thetaftFinal}
\end{align}
where the hyperparameter $\lambda$ controls the regularization strength.
Intuitively, $F_i$ corresponds to the local importance of parameter $\theta_i$.
The regularization, $\|\theta - \thetassl\|_{F(\thetassl)}^2$, constrains
important parameters, i.e.\ those with high $F$ values, to stay close to their
initial pre-trained SSL values, while allowing less important parameters to vary
more freely. In our experiments, we also consider the naive setting, where we replace
the \gls{fim} regulariztion in Eq \eqref{eqn:thetaftFinal} with a
quadratic L2 penalty, $\|\theta - \thetassl\|_{2}^2$.

We focus on DINO \citep{DBLP:journals/corr/abs-2104-14294} due to the high
performant nature of the model, and show that it can be recast in the
probabilistic framework described above. DINO uses a teacher network to produce
pseudo-labels for the cross-entropy loss and a student network is used to
predict those labels. As is common in SSL, both networks consist of a backbone
network $f_\theta$ and a projection head network $g_\theta$. Given an image,
$x$, drawn from an augmentation distribution
$\{x_{1}, x_{2}\} \sim \mathcal{A}(x)$, the student network,
$s(x_{1}; \theta) = g_\theta(f_\theta(x_{1}))$, applies these networks
sequentially while the teacher network inserts a centering operation in between
$t(x_{2}; \theta) = g_\theta(\text{centering}(f_\theta(x_{2})))$. While the
student network parameters are updated to minimize the cross-entropy between the
teacher and student outputs, the teacher network parameters are set to the
exponentially moving average \citep{polyak1992acceleration} of the student
parameters. Therefore, the student network can be interpreted as a probabilistic
model $p_\theta(y = c | x) = \text{softmax}(s(x; \theta))_c$,
while the teacher network provides pseudo-labels for the the dataset $\calD_\text{SSL} = \{(x_i, t(x_i; \theta))\}_{i=1}^N$.

\section{Experiments}\label{sec:experiments}

Given a pre-trained ImageNet1k DINO model, we compute the \gls{fim} using the
first 10000 images of ImageNet1k. We consider three downstream tasks in this
work: (i) Waterbirds \citep{DBLP:journals/corr/abs-1911-08731} (ii) Celeb-A
\citep{DBLP:conf/iccv/LiuLWT15} and (iii) CIFAR10
\citep{Krizhevsky09learningmultiple}. The full fine-tuning procedure is
described in detail in Appendix \ref{tab:cifar10_hps}. Celeb-A and Waterbirds
are datasets with known (or constructed) biases and are used for analyzing
worst sub-group performance \citep{DBLP:journals/corr/abs-1911-08731}. By
interpreting the deterioration of worst sub-group performance as a catastrophic
forgetting \citep{mccloskey1989catastrophic} event for a robust SSL model
(trained on large scale data), we can leverage \gls{cl} techniques such as \gls{ewc}
to minimize this performance penalty.
\paragraph{FIM analysis}
\begin{figure}
    \begin{center}
            \begin{minipage}{0.495\textwidth}
                \begin{center}
                    \centering
                    \includegraphics[width=\linewidth]{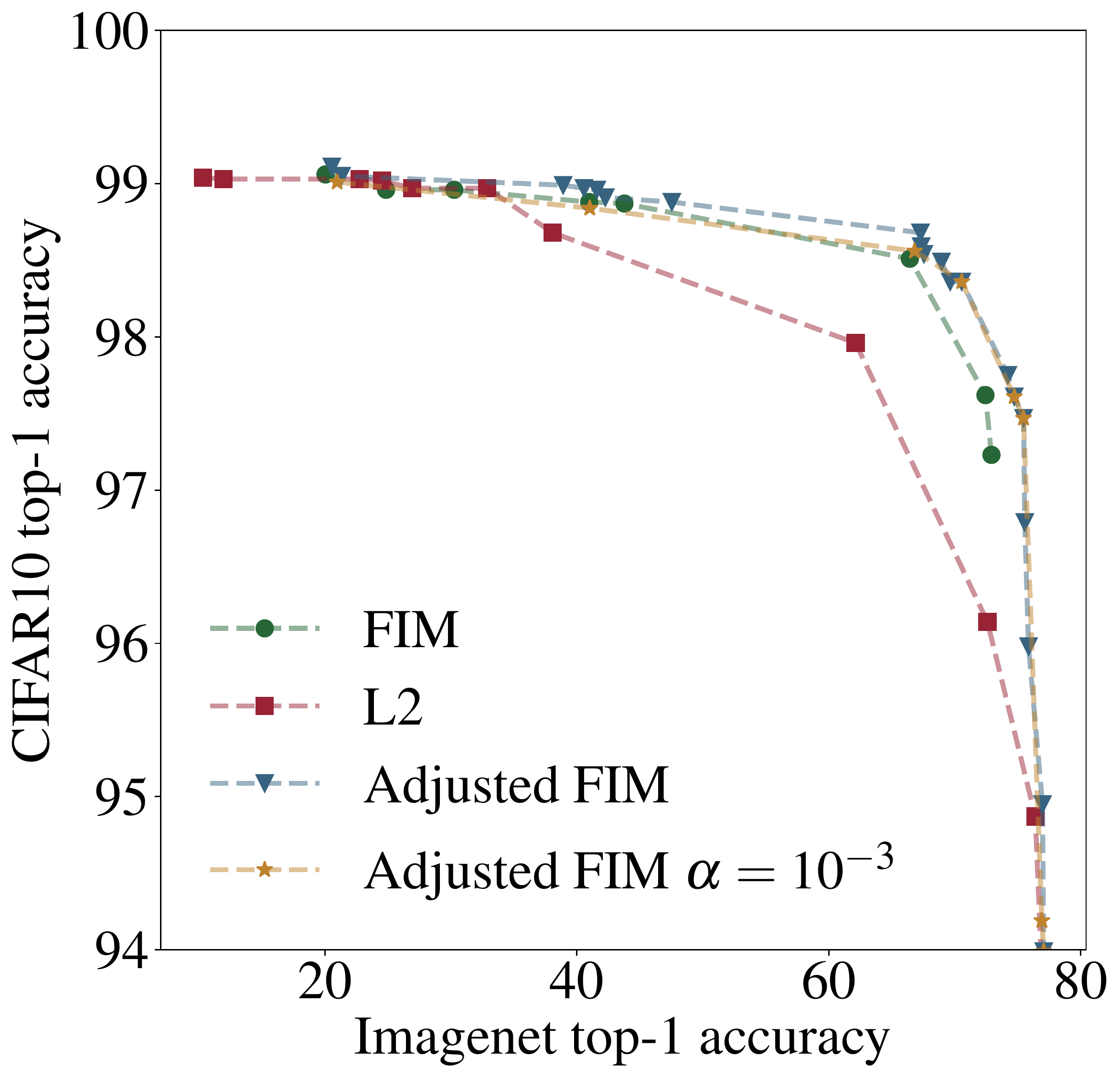}
                \end{center}
                \setlength{\belowcaptionskip}{-10pt}
                \caption{\small Pareto fronts of different regularization methods when fine-tuning a DINO ViT-B/16 pretrained on ImageNet to CIFAR10.}\label{fig:cifar10_pareto}
            \end{minipage}
            \hfill
            \begin{minipage}{0.495\textwidth}
                \begin{center}
                    \centering
                    \includegraphics[width=\linewidth]{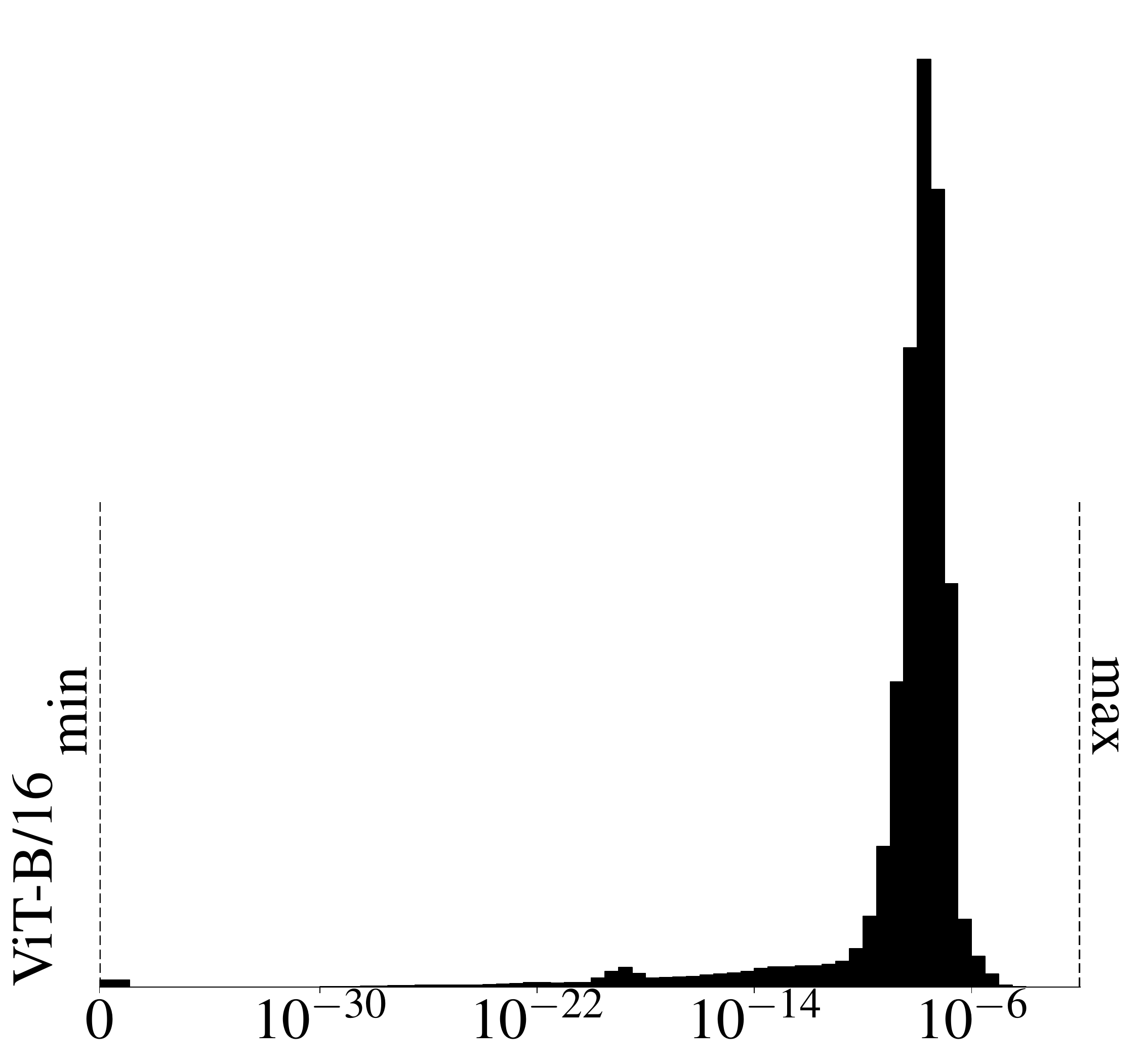}
                    \phantom{Hello}
                \end{center}
                \setlength{\belowcaptionskip}{-10pt}
                \caption{\small FIM value distributions on ViT-B/16}\label{fig:fim_distribution}
                    \phantom{Hello}
                    \phantom{Hello}
                    \phantom{Hello}
                    \phantom{Hello}
                    \phantom{Hello}
                    \phantom{Hello}
                    \phantom{Hello}
                    \phantom{Hello}
                    \phantom{Hello}
                    \phantom{Hello}
                    \phantom{Hello}
                    \phantom{Hello}
                    \phantom{Hello}
                    \phantom{Hello}
                    \phantom{Hello}
                    \phantom{Hello}
            \end{minipage}
    \end{center}
\end{figure}

We start by validating the expected properties of \gls{fim} by fine-tuning the
pre-trained DINO ViT-B/16 ImageNet1k model on the CIFAR10 dataset. We vary the
learning rate and the \gls{fim} regularization weight\footnote{For the adjusted
FIM experiments we also vary $\alpha$.}. After training, we evaluate the
performance by computing: a) top-1 accuracy on the CIFAR10 dataset, and b) top-1
accuracy on ImageNet1k dataset. To evaluate the reverse transfer performance
from CIFAR10 to ImageNet1k we attach the ImageNet1k classification head from the
\emph{pre-trained model} to the \emph{fine-tuned backbone}.

Figure~\ref{fig:cifar10_pareto} shows the cross-task Pareto fronts for each fine-tuning method.
We observe that as the regularization weight is increased, both methods improve in terms
of their performance on ImageNet1k. However, since \gls{fim} regularizes parameters preferentially,
its performance on CIFAR10 does not decrease in contrast to the naive $L_2$ regularization approach discussed in Section \ref{sec:methodology}.

While the $L_2$ regularization approach manages to fully recover 77\% top-1 accuracy on
ImageNet1k this is not the case for \gls{fim} regularization, even for very high
regularization weights $\lambda$, which saturates at around 73\% (Figure~\ref{fig:cifar10_pareto}).

We attribute this to the fact that a significant proportion of
parameters in the ViT-B/16 have low values (see the distribution of
FIM values in Figure~\ref{fig:fim_distribution} and refer to Appendix \ref{sec:fimDist} for
a more detailed breakdown).
By decomposing the \gls{fim} into a layer-wise distributional plot
(Figure \ref{fig:fimViTB16}) we observe that the first attention layer has a
large proportion of very small values within the range of $10^{-30}$.
In fact, some of the FIM values are 0, making $\|\cdot\|_F$ a pseudo-norm, i.e.\ minimizing $\|\cdot\|_F$ does not
imply fully recovering all of the parameters. In other words, $F$ is poorly-conditioned,
and parameters with small FIM values are allowed to vary freely making
full recovery of the model impossible.
To resolve this we rescale $F$:
\begin{align}
    F_\alpha = (1-\alpha)F + \alpha\bar{F}, \label{eqn:adjusted_fim}
\end{align}
where $\bar{F}$ is the mean of $F$ and $\alpha \in [0, 1]$ is an additional hyperparameter.
This rescaling pushes $\|\cdot\|_F$ closer to $\|\cdot\|_2$, which is not ill-conditioned. \emph{Adjusted FIM} in Figure~\ref{fig:cifar10_pareto} sweeps the Pareto front with varying $\alpha$ whereas \emph{Adjusted FIM} $\alpha=10^{-3}$ sweeps the Pareto front with a fixed $\alpha$ -- both strategies for rectification produce a Pareto front that improves on $L_2$ and the unadjusted FIM regularization.

\paragraph{Robustness analysis}

\begin{table}[t]
\centering
\caption{\small
  Results of fine-tuning a pre-trained DINO model.
  We report results for our naive \emph{$L_{2}$}
regularization method $\|\theta - \thetassl\|_{2}^2$, \gls{ewc} \emph{FIM} and
the \emph{Adjusted FIM} ($\alpha=10^{-3}$) regularized variants.
We compare against a competitively tuned \gls{erm} baseline \citep{DBLP:books/sp/95/V1995}, finetuned from the same pre-trained DINO model -- this differs from the pre-trained supervised initialization of \citet{DBLP:journals/corr/abs-1911-08731}.
For more details and hyperparameters see Appendix \ref{sec:hps}.
Given a validation set \emph{without group labels}, we choose hyperparameters with the highest validation top-1 accuracy. Results are
reported on the worst sub-group test split over 5 trials (mean $\pm$ 1-std).}
\small
\begin{tabular}{llllll}
\toprule
\multirow{2}*{ } & \multirow{2}*{\textbf{Finetuning method}} & \textit{Waterbirds} & \textit{CelebA} \\
\cmidrule[0.625pt](l){5-6}
{} & {} & \multicolumn{1}{c}{\textit{Test WGA}} & \multicolumn{1}{c}{\textit{Test WGA}}  \\
\toprule
\multirow{4}{*}{\rotatebox[origin=c]{90}{\textit{ViT-B/16}}}
                                   & ERM                                            & $70.53 (\pm 2.07)$                                     & $41.22 (\pm 1.86)$                                   \\
\cmidrule(l){2-6}
                                   & FIM (ours)                                     & $75.39 (\pm 1.38)$                                     & $43.22 (\pm 1.49)$                                   \\
                                   & Adjusted FIM (ours)                            & $71.00 (\pm 3.55)$                                     & $43.88 (\pm 2.52)$                                                                      \\
                                   & $L_2$ (ours)                                   & $72.43 (\pm 1.30)$                                     & $38.89 (\pm 1.62)$\\
\toprule
\multirow{4}{*}{\rotatebox[origin=c]{90}{\textit{ResNet50}}} & ERM                  & $62.55 (\pm 1.73)$ & $40.55 (\pm 1.47)$                             \\
\cmidrule(l){2-6}
                                   & FIM (ours)                                     & $60.90 (\pm 0.54)$                                     & $41.85 (\pm 1.40)$                                   \\
                                   & Adjusted FIM (ours)                            & $60.29  (\pm 1.63)$                                     & $41.85 (\pm 1.70)$                                   \\
                                   & $L_2$ (ours)                                   & $39.84 (\pm 2.26)$                                     & $38.15 (\pm 0.85)$                                   \\
\bottomrule
\end{tabular}
\label{tab:results}
\end{table}

We use the proposed regularization to finetune DINO models on the biased
Waterbirds \citep{DBLP:journals/corr/abs-1911-08731} and CelebA
\citep{DBLP:conf/iccv/LiuLWT15} datasets. We fine-tune all of the mentioned
methods with varying hyperparameters, extracted from a 20 trial random
hyper-parameter sweep, evaluated over 5 replicates. We evaluate a simplified
scenario where we assume we don't have access to group information and pick the
validation hyper-parameters that maximize the total top-1 accuracy. This
parallels the \gls{erm} evaluation from
\cite{DBLP:journals/corr/abs-1911-08731}. We emphasize that most other methods for
improving worst sub-group performance, such as~\cite{DBLP:journals/corr/abs-1911-08731, DBLP:conf/icml/LiuHCRKSLF21, DBLP:journals/corr/abs-2204-02937},
assume access to group labels in the validation set and therefore are not a
suitable comparison.

We find that FIM based regularization techniques outperform standard finetuning
in all but one scenario (ResNet50 on Waterbirds), demonstrating the
utility of regularizing towards previously learned SSL representations.
With ViT-B/16 we observe a 5\% improvement on Waterbirds
and 2\% improvement on CelebA compared to our baseline ERM model. With ResNet50
this regularization improves upon ERM by approximately 1\% on CelebA, but performs
worse on the Waterbirds dataset.

\section{Conclusion}

In this work we present a novel treatment of \gls{ssl} finetuning under the lens
of a Bayesian continual learning framework. We empirically demonstrate that
constraining the fine-tuned model posterior towards the original (label) bias
free pre-trained model posterior improves performance for the worst sub groups
in Celeb-A and Waterbirds. Future work will explore how other \gls{cl}
techniques could be leveraged to further improve \gls{ssl} finetuning
procedures.

\section{Acknowledgements}

The authors would like to thank the following people for their help throughout 
the process of writing this paper, in alphabetical order: Barry-John Theobald 
and Miguel Sarabia del Castillo. 
Additionally, we thank Li Li, Mubarak Seyed Ibrahim, Okan Akalin, and the wider 
Apple infrastructure team for assistance with developing scalable, fault tolerant code.

\bibliography{libraries/main}
\bibliographystyle{templates/iclr2021/iclr2021_conference}

\clearpage
\appendix

\section{Experiment details} \label{sec:hps}

\begin{table}[h]
  \centering
    \small
    \caption{Hyperparameters for finetuning on CIFAR10}\label{tab:cifar10_hps}
    \begin{tabular}{cccc}
        \toprule
        \multirow{2}*{\textbf{Hyperparameter}} & \multicolumn{3}{c}{\textbf{Finetuning method}} \\
        \cmidrule[0.5pt](l){2-4}
        & FIM & adjusted FIM & $L_2$ \\
        \toprule
        learning rate & $\{0.3, 1, 3\}\times 10^{-5}$ & $\{1, 3\}\times 10^{-5}$ & $\{0.3 , 1, 3 \}\times 10^{-5}$ \\
        \cmidrule(l){1-4}
        regularization weight & $0, 10^{\{4,5,6,7,8,9,10\}}$ & $10^{\{4,5,6,7,8,9\}}$ & $0, 10^{\{-4,-3,-2,-1,0,1, 2\}}$ \\
        \cmidrule(l){1-4}
        $\alpha$ & --- & $10^{\{-8,-7,-6,-4,-3,-2,-1\}}$ & --- \\
        \cmidrule(l){1-4}
        batch size & 4032 & 4032 & 4032 \\
        \cmidrule(l){1-4}
        epochs & 120 & 120 & 120 \\
        \cmidrule(l){1-4}
        warmup epochs & 20 & 20 & 20 \\
        \cmidrule(l){1-4}
        drop path rate & 0 & 0 & 0 \\
        \cmidrule(l){1-4}
        optimizer & adamw & adamw & adamw \\
        \cmidrule(l){1-4}
        learning rate schedule & cosine & cosine & cosine \\
        \cmidrule(l){1-4}
        learning rate scaling value & 256 & 256 & 256 \\
        \cmidrule(l){1-4}
        weight decay & 0.3 & 0.3 & 0.3 \\
        \cmidrule(l){1-4}
        weight decay end & 0.7 & 0.7 & 0.7 \\
        \cmidrule(l){1-4}
        augmentation & randaug & randaug & randaug \\
        \cmidrule(l){1-4}
        aug multiplicity & 2 & 2 & 2 \\
        \bottomrule
    \end{tabular}
\end{table}
CIFAR10 experiments in Section~\ref{sec:experiments} were run with hyperparameters in Table~\ref{tab:cifar10_hps}. Each hyperparameter combination was run
once and its performance was evaluated at the end of training on CIFAR10 and
Imagenet1k test sets. As mentioned before, to evaluate the performance of the
fintuned backbone on Imagenet1k, we attached an Imagenet1k classifier head which
was trained before the backbone was finetuned. Figure~\ref{fig:cifar10_pareto} was
produced by computing the convex hull over the (CIFAR10 top-1 accuracy, Imagenet1k top-1 accuracy)
data points for each group of runs.
\begin{table}[h]
  \centering
    \small
    \caption{Hyperparameter sweep for finetuning ResNet50 on Waterbirds}\label{tab:waterbirds_r50_hps}
    \begin{tabular}{ccccc}
        \toprule
        \multirow{2}*{\textbf{Hyperparameter}} & \multicolumn{3}{c}{\textbf{Finetuning method}} \\
        \cmidrule[0.5pt](l){2-4}
        & FIM & adjusted FIM & $L_2$ & ERM \\
        \toprule
        learning rate & \multicolumn{4}{c}{$10^{-2, -3, -4}$} \\
        \cmidrule(l){1-5}
        regularization weight & $10^{\{5,6,7\}}$ & $10^{\{5,6,7\}}$ & $10^{\{-1,0,1\}}$ & --- \\
        \cmidrule(l){1-5}
        $\alpha$ & --- & $10^{-3}$ & --- & --- \\
        \cmidrule(l){1-5}
        batch size & \multicolumn{4}{c}{128, 256} \\
        \cmidrule(l){1-5}
        epochs & \multicolumn{4}{c}{120} \\
        \cmidrule(l){1-5}
        warmup epochs & \multicolumn{4}{c}{0} \\
        \cmidrule(l){1-5}
        drop path rate & \multicolumn{4}{c}{0} \\
        \cmidrule(l){1-5}
        optimizer & \multicolumn{4}{c}{SGD(momentum=0.9)} \\
        \cmidrule(l){1-5}
        learning rate schedule & \multicolumn{4}{c}{constant} \\
        \cmidrule(l){1-5}
        learning rate scaling value & \multicolumn{4}{c}{1.0} \\
        \cmidrule(l){1-5}
        weight decay & \multicolumn{4}{c}{$\{0, 10^{-4}, 1.5 \times 10^{-6}\}$} \\
        \cmidrule(l){1-5}
        weight decay end & \multicolumn{4}{c}{---} \\
        \cmidrule(l){1-5}
        augmentation & \multicolumn{4}{c}{imagenet, crop\_scale=(0.2, 1.0)} \\
        \cmidrule(l){1-5}
        aug multiplicity & \multicolumn{4}{c}{1} \\
        \bottomrule
    \end{tabular}
  \end{table}
\begin{table}[h]
  \centering
    \small
    \caption{Hyperparameter sweep for finetuning ViT-B/16 on Waterbirds}\label{tab:waterbirds_vitb_hps}
    \begin{tabular}{ccccc}
        \toprule
        \multirow{2}*{\textbf{Hyperparameter}} & \multicolumn{3}{c}{\textbf{Finetuning method}} \\
        \cmidrule[0.5pt](l){2-4}
        & FIM & adjusted FIM & $L_2$ & ERM \\
        \toprule
        learning rate & \multicolumn{4}{c}{$10^{-6, -5, -4}$} \\
        \cmidrule(l){1-5}
        regularization weight & $10^{\{5,6,7\}}$ & $10^{\{5,6,7\}}$ & $10^{\{-1,0,1\}}$ & --- \\
        \cmidrule(l){1-5}
        $\alpha$ & --- & $10^{-3}$ & --- & --- \\
        \cmidrule(l){1-5}
        batch size & \multicolumn{4}{c}{128, 256} \\
        \cmidrule(l){1-5}
        epochs & \multicolumn{4}{c}{120} \\
        \cmidrule(l){1-5}
        warmup epochs & \multicolumn{4}{c}{0} \\
        \cmidrule(l){1-5}
        drop path rate & \multicolumn{4}{c}{0} \\
        \cmidrule(l){1-5}
        optimizer & \multicolumn{4}{c}{AdamW($\beta_{1}=0.9, \beta_{2}=0.95$)} \\
        \cmidrule(l){1-5}
        learning rate schedule & \multicolumn{4}{c}{cosine} \\
        \cmidrule(l){1-5}
        learning rate scaling value & \multicolumn{4}{c}{1.0} \\
        \cmidrule(l){1-5}
        weight decay & \multicolumn{4}{c}{$\{0, 10^{\{-3, -2, -1\}}\}$} \\
        \cmidrule(l){1-5}
        weight decay end & \multicolumn{4}{c}{---} \\
        \cmidrule(l){1-5}
        augmentation & \multicolumn{4}{c}{imagenet, crop\_scale=(0.2, 1.0)} \\
        \cmidrule(l){1-5}
        aug multiplicity & \multicolumn{4}{c}{1} \\
        \bottomrule
    \end{tabular}
  \end{table}

  Waterbirds experiments in Section~\ref{sec:experiments} were run with hyperparameters in Table~\ref{tab:waterbirds_r50_hps} and Table~\ref{tab:waterbirds_vitb_hps}. 18 combinations were randomly sampled without replacement and each sampled hyperparameter combination was run five times. We select the best performing validation hyper-parameter set for the top-1 metric (using \emph{no group labels}) and use that to report the test worst sub-group accuracy for all methods.
\begin{table}[h]
  \centering
    \small
    \caption{Hyperparameter sweep for finetuning ResNet50 on Celeb-A}\label{tab:celeba_r50_hps}
    \begin{tabular}{ccccc}
        \toprule
        \multirow{2}*{\textbf{Hyperparameter}} & \multicolumn{3}{c}{\textbf{Finetuning method}} \\
        \cmidrule[0.5pt](l){2-4}
        & FIM & adjusted FIM & $L_2$ & ERM \\
        \toprule
        learning rate & \multicolumn{4}{c}{$10^{-2, -3, -4}$} \\
        \cmidrule(l){1-5}
        regularization weight & $10^{\{5,6,7\}}$ & $10^{\{5,6,7\}}$ & $10^{\{-3,-2,-1\}}$ & --- \\
        \cmidrule(l){1-5}
        $\alpha$ & --- & $10^{-3}$ & --- & --- \\
        \cmidrule(l){1-5}
        batch size & \multicolumn{4}{c}{2048} \\
        \cmidrule(l){1-5}
        epochs & \multicolumn{4}{c}{300} \\
        \cmidrule(l){1-5}
        warmup epochs & \multicolumn{4}{c}{0} \\
        \cmidrule(l){1-5}
        drop path rate & \multicolumn{4}{c}{0} \\
        \cmidrule(l){1-5}
        optimizer & \multicolumn{4}{c}{LARS(momentum=0.9)} \\
        \cmidrule(l){1-5}
        learning rate schedule & \multicolumn{4}{c}{cosine} \\
        \cmidrule(l){1-5}
        learning rate scaling value & \multicolumn{4}{c}{256.0} \\
        \cmidrule(l){1-5}
        weight decay & \multicolumn{4}{c}{$\{0, 10^{-4}, 1.5 \times 10^{-6}\}$} \\
        \cmidrule(l){1-5}
        weight decay end & \multicolumn{4}{c}{---} \\
        \cmidrule(l){1-5}
        augmentation & \multicolumn{4}{c}{imagenet, crop\_scale=(0.2, 1.0)} \\
        \cmidrule(l){1-5}
        aug multiplicity & \multicolumn{4}{c}{1} \\
        \bottomrule
    \end{tabular}
\end{table}
\begin{table}[h]
  \centering
    \small
    \caption{Hyperparameter sweep for finetuning ViT-B/16 on Celeb-A}\label{tab:celeba_vitb_hps}
    \begin{tabular}{ccccc}
        \toprule
        \multirow{2}*{\textbf{Hyperparameter}} & \multicolumn{3}{c}{\textbf{Finetuning method}} \\
        \cmidrule[0.5pt](l){2-4}
        & FIM & adjusted FIM & $L_2$ & ERM \\
        \toprule
        learning rate & \multicolumn{4}{c}{$10^{-6, -5, -4}$} \\
        \cmidrule(l){1-5}
        regularization weight & $10^{\{5,6,7\}}$ & $10^{\{5,6,7\}}$ & $10^{\{-1,0,1\}}$ & --- \\
        \cmidrule(l){1-5}
        $\alpha$ & --- & $10^{-3}$ & --- & --- \\
        \cmidrule(l){1-5}
        batch size & \multicolumn{4}{c}{2048, 4096} \\
        \cmidrule(l){1-5}
        epochs & \multicolumn{4}{c}{60} \\
        \cmidrule(l){1-5}
        warmup epochs & \multicolumn{4}{c}{0} \\
        \cmidrule(l){1-5}
        drop path rate & \multicolumn{4}{c}{0} \\
        \cmidrule(l){1-5}
        optimizer & \multicolumn{4}{c}{AdamW($\beta_{1}=0.9, \beta_{2}=0.95$)} \\
        \cmidrule(l){1-5}
        learning rate schedule & \multicolumn{4}{c}{cosine} \\
        \cmidrule(l){1-5}
        learning rate scaling value & \multicolumn{4}{c}{256.0} \\
        \cmidrule(l){1-5}
        weight decay & \multicolumn{4}{c}{$\{0, 10^{\{-3, -2, -1\}}\}$} \\
        \cmidrule(l){1-5}
        weight decay end & \multicolumn{4}{c}{---} \\
        \cmidrule(l){1-5}
        augmentation & \multicolumn{4}{c}{imagenet, crop\_scale=(0.2, 1.0)} \\
        \cmidrule(l){1-5}
        aug multiplicity & \multicolumn{4}{c}{1} \\
        \bottomrule
    \end{tabular}
  \end{table}

  Celeb-A experiments in Section~\ref{sec:experiments} were run with hyperparameters in Table~\ref{tab:celeba_r50_hps} and Table~\ref{tab:celeba_vitb_hps}. 18 combinations were randomly sampled without replacement and each sampled hyperparameter combination was run five times. We select the best performing validation hyper-parameter set for the top-1 metric (using \emph{no group labels}) and use that to report the test worst sub-group accuracy for all methods.

\section{FIM distributions}\label{sec:fimDist}

\begin{figure}[h]
    \begin{center}
        \scalebox{0.9}{\parbox{1.0\linewidth}{%
            \begin{minipage}{0.49\textwidth}
                \begin{center}
                    \centering
                    \includegraphics[width=\linewidth]{figures/fim/vitb16/fim.pdf}
                \end{center}
                \caption{Overall \gls{fim} distribution of ViT-B/16}
            \end{minipage}
            \hfill
            \begin{minipage}{0.49\textwidth}
                \begin{center}
                    \centering
                    \includegraphics[width=\linewidth]{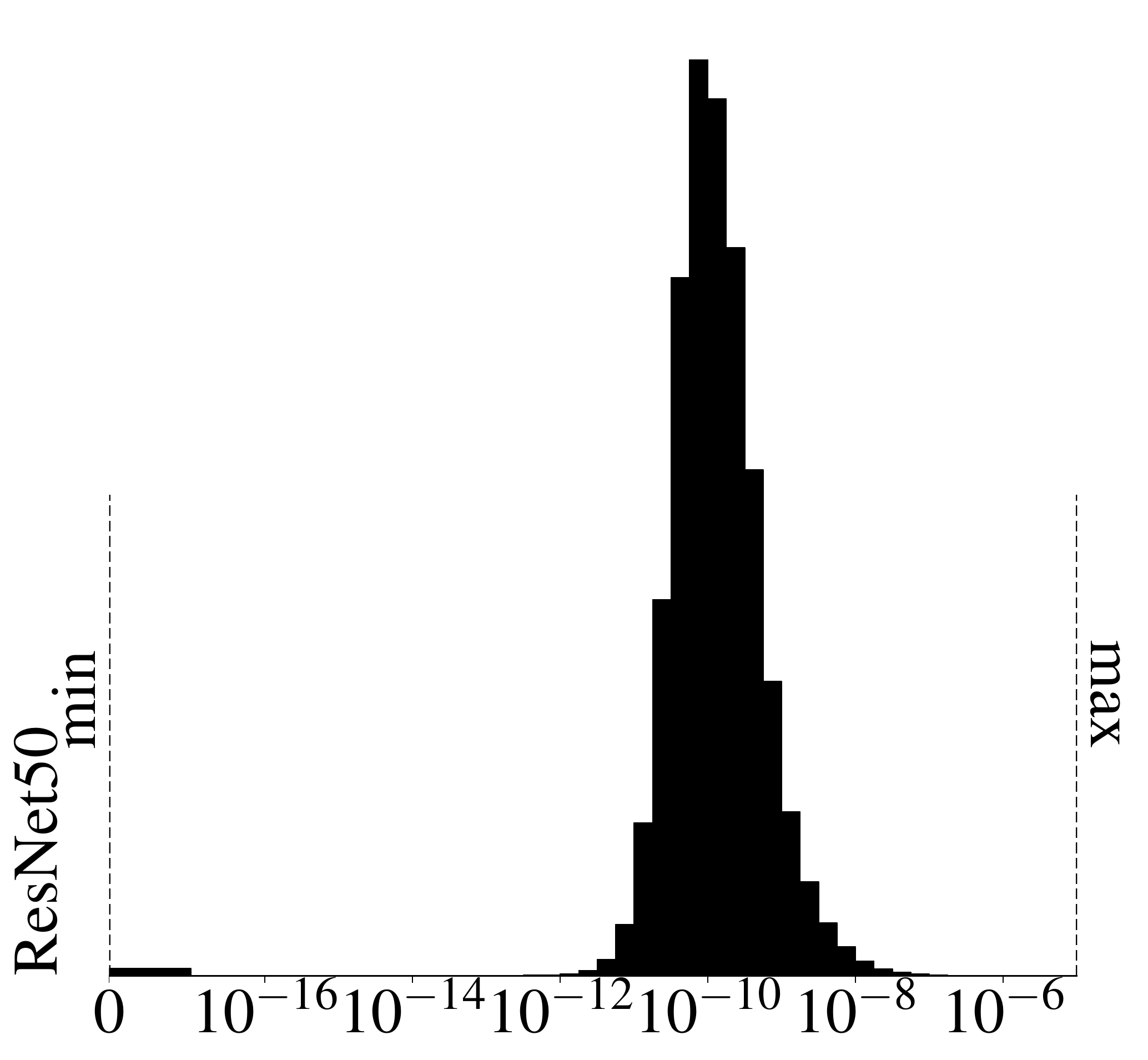}
                \end{center}
                \caption{Overall \gls{fim} distribution of ResNet50}
            \end{minipage}
        }}
    \end{center}
\end{figure}

\begin{figure}
    \begin{center}
        \scalebox{0.9}{\parbox{1.0\linewidth}{%
            \begin{minipage}{0.49\textwidth}
                \begin{center}
                    \centering
                    \includegraphics[width=\linewidth]{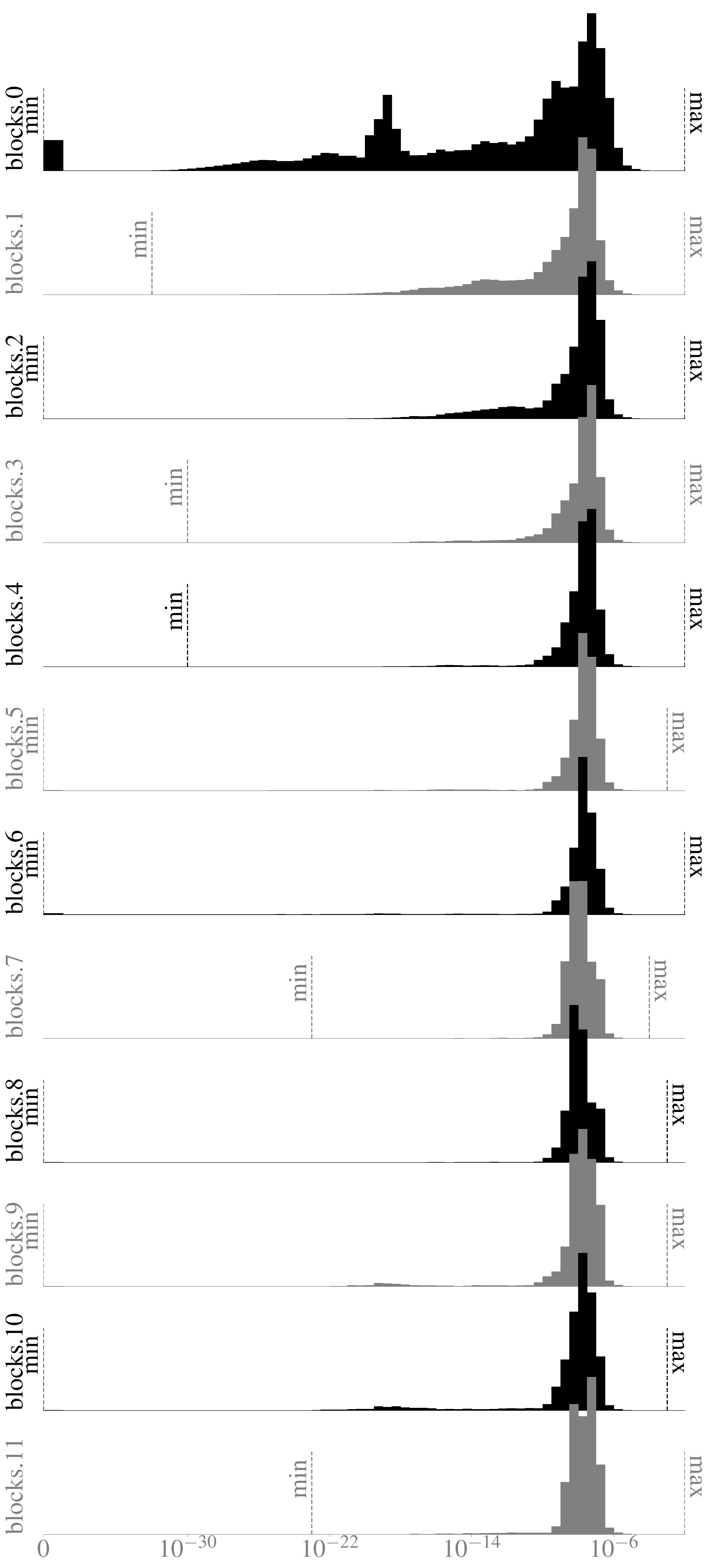}
                \end{center}
                \caption{\gls{fim} distribution of outermost layers of ViT-B/16. Note that the earlier blocks have much wider spread of \gls{fim} values.}\label{fig:fimViTB16}
            \end{minipage}
            \hfill
            \begin{minipage}{0.49\textwidth}
                \begin{center}
                    \centering
                    \includegraphics[width=\linewidth]{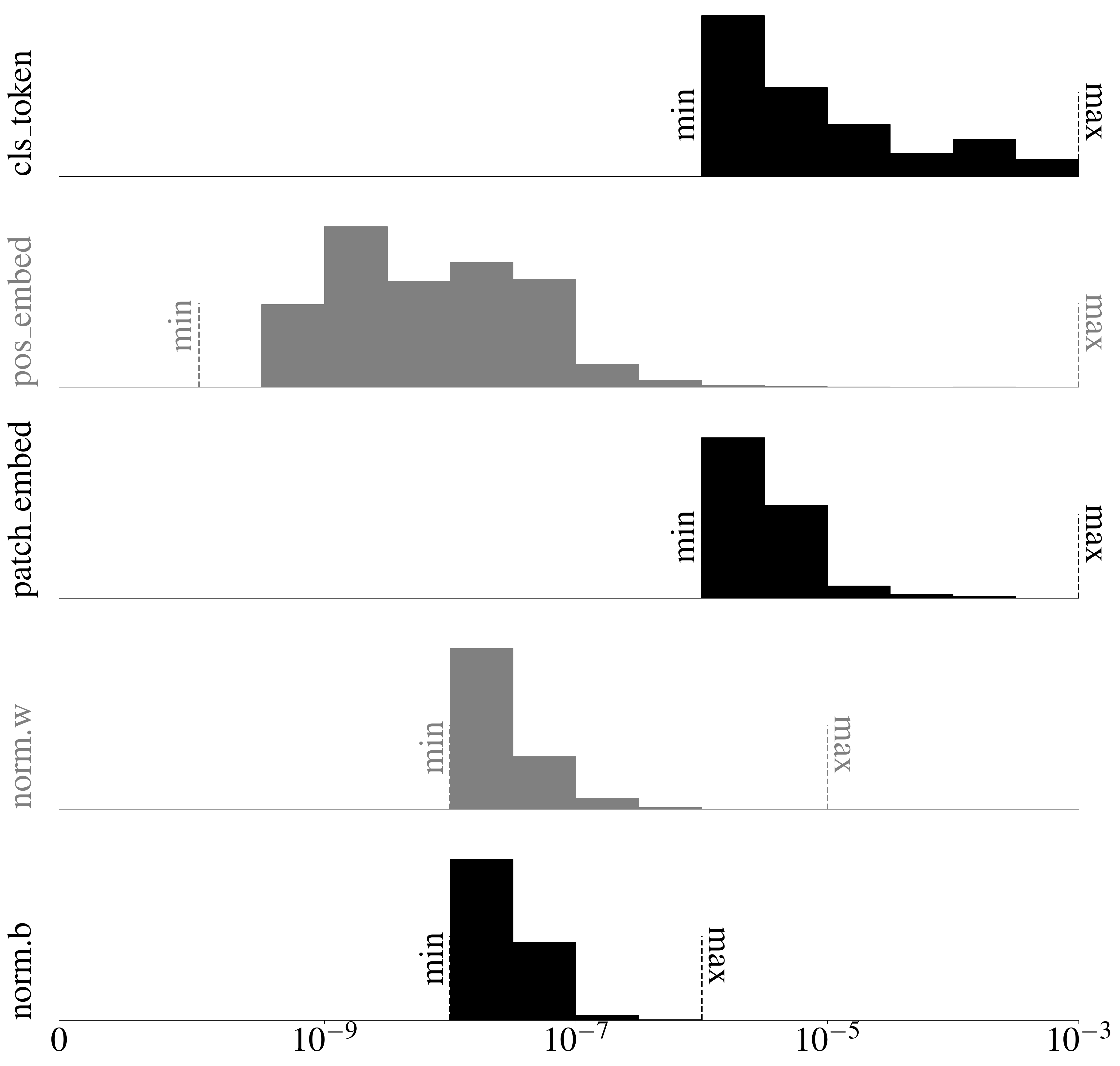}
                \end{center}
                \caption{\gls{fim} distribution of the non-block layers of ViT-B/16.}
                \begin{center}
                    \centering
                    \includegraphics[width=\linewidth]{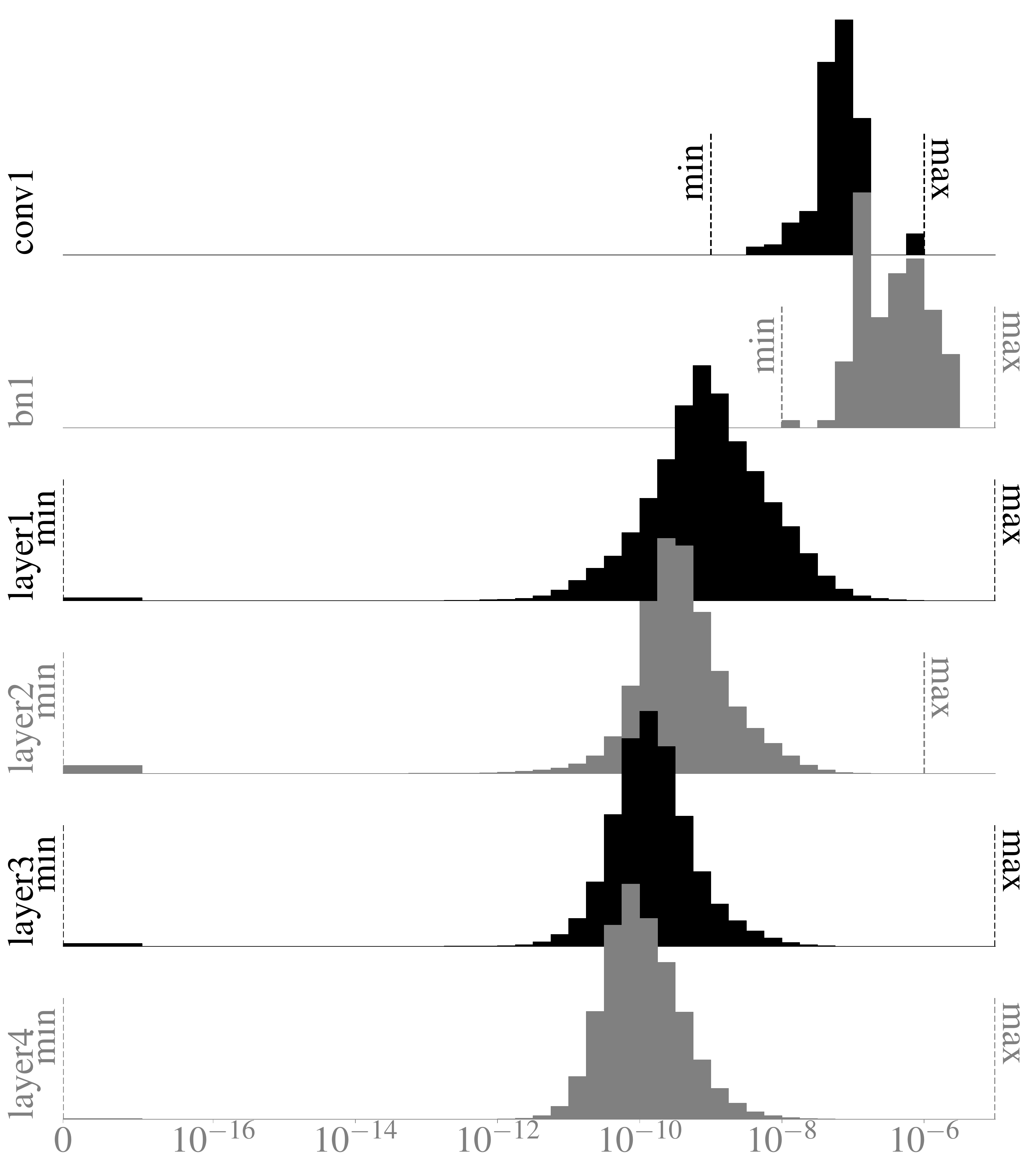}
                \end{center}
                \caption{\gls{fim} distribution of outermost layers of ResNet50. Note that while the values seem better behaved than in the case of ViT-B/16, all blocks still have some parameters with 0 \gls{fim} value.}
            \end{minipage}
        }}
    \end{center}
\end{figure}

\end{document}